\documentclass{article}

\usepackage[preprint]{neurips_2026}

\usepackage[utf8]{inputenc} 
\usepackage[T1]{fontenc}    
\usepackage{hyperref}       
\usepackage{url}            
\usepackage{booktabs}       
\usepackage{amsfonts}       
\usepackage{nicefrac}       
\usepackage{microtype}      
\usepackage{xcolor}         
\usepackage{tabularx}
\usepackage{array}

\usepackage{tikz}
\usetikzlibrary{positioning, arrows.meta} 
\usepackage{adjustbox}
\usepackage{subcaption}
\usepackage{amsmath}

\title{Private Vertical Federated Inference for Time-Series}

\author{%
  Lucas Fenaux\thanks{Equal contribution} \\
  University of Waterloo\\
  Waterloo, Canada \\
  \texttt{lucas.fenaux@uwaterloo.ca} \\
  \And
  Larris Xie\footnotemark[1] \\
  University of Waterloo\\
  Waterloo, Canada \\
  \texttt{larris.xie@uwaterloo.ca} \\
  \AND
  Aditya Bang \\
  University of Waterloo\\
  Waterloo, Canada \\
  \texttt{a3bang@uwaterloo.ca} \\
  \And
  Alex Zhang \\
  University of Waterloo\\
  Waterloo, Canada \\
  \texttt{alex.zhang2@uwaterloo.ca} \\
  \And
  Kevin Wilson \\
  Borealis AI\\
  Toronto, Canada \\
  \texttt{} \\
  \And
  Florian Kerschbaum \\
  University of Waterloo\\
  Waterloo, Canada \\
  \texttt{florian.kerschbaum@uwaterloo.ca} \\
}

\begin{document}

\maketitle

\begin{abstract}
Institutions may benefit from collaborative inference on time-series data. In settings where privacy is necessary, multi-party computation (MPC) is a straightforward approach to providing strong guarantees, yet it remains prohibitively expensive and scales poorly with modern transformer architectures. Vertical Federated Learning (VFL) offers efficiency but suffers from privacy leakage at the embedding level, and securing the entire VFL model head via MPC remains prohibitively slow and communication-heavy for larger models. To enable practical, secure inference at scale, we propose \textbf{Public/Private Hybrid Head-VFL (PPHH-VFL)}. This hybrid architecture splits the model head into an efficient plaintext public head and a secure, lightweight MPC private head. By applying adversarial training to the public embeddings, we mitigate privacy leakage; concurrently, the small private head securely preserves the flow of sensitive information needed for high downstream utility. Empirical evaluations on models ranging up to 86 million parameters demonstrate that PPHH-VFL accelerates inference by up to six orders of magnitude compared to end-to-end MPC. Compared to a standard VFL+MPC baseline, our approach scales significantly better, achieving a speedup of up to 44.4x in WAN and a 91.2x reduction in communication costs (dropping from 1.7 GB to 19 MB per batch), while simultaneously improving downstream classification accuracy by 2.50\% and regression RMSE by 40.7\%.
\end{abstract}

\section{Introduction}

Time-series data underpins decision-making in several domains where privacy constraints are intrinsic rather than optional. Credit card fraud detection systems, clinical monitoring streams \cite{tang2023ihvfl, shankar2025share}, and user-level behavioral telemetry \cite{shankar2025share} are examples of use cases where collaboration between data-holding institutions and access to large models that can handle long temporal context are extremely valuable. However, these use cases, which could strongly benefit from collaborative inference and access to large models, are also the ones where data privacy is often of the utmost concern.

In theory, multi-party computation (MPC) protocols enable collaborative, private inference over time-series data while protecting users' privacy \cite{ben2019completeness}, e.g., Yao's garbled circuits \cite{demmlerABYFrameworkEfficient2015} or additive secret sharing \cite{knottCrypTenSecureMultiParty2021}. However, their runtime scales poorly as model size and input length increase, especially in higher latency settings. This poor scaling is particularly evident in transformer models \cite{vaswani2017attention}, the premier model architecture for tasks that rely on time-series data \cite{li2022mpcformer, wang2022characterization}. Attention layers are the main drivers of overhead during inference. It has been well established in the literature that activation functions are the bottleneck in MPC-based secure inference, taking up to 93\% of the inference time~\cite{ding2023east, garimellaSisyphusCautionaryTale2021,hussainCOINNCryptoML2021,mishraDelphiCryptographicInference2020,CryptoNASProceedings34th}, as they require multiple rounds of communication between clients. Attention specifically can take up to 80\% of the overall inference runtime for transformer models, with the softmax operation itself taking 50\% to 67.8\% of the overall runtime \cite{li2022mpcformer, wang2022characterization}.

In such scenarios, Federated Learning (FL) \cite{mcmahan2017communication} is a useful alternative to MPC that is more efficient, but less privacy-protecting. It enables distributed machine learning in which multiple clients collaboratively train or infer a model without directly sharing their data. 
We consider the following scenario: a data sample consists of a time sequence of features, and the time steps within that sequence are split among multiple clients. For example, a data sample is the time sequence of credit card transactions made by a user with multiple credit cards across different banks. Each bank knows the transactions made by the user with its respective credit card, but not the others'. We can use Vertical Federated Learning (VFL) \cite{yang2019federated}, a variant of FL in which feature samples are split across clients. Unlike traditional VFL, where the feature vector at each time step is split among clients \cite{shankar2025share, yang2025mvfl}, we propose splitting the time steps across clients. To the best of our knowledge, this application of assigning different time segments to different clients introduces a new perspective on the VFL model.

In the VFL model, each client computes a local embedding of its input using its own private local model. Then, they send their embeddings to the server, which aggregates them and computes the final part of the model, the head, and extracts the model's prediction. The problem with this approach is that embeddings are not inherently private, and privacy attacks can be carried out with them at inference time \cite{morris2023text, sun2022label}. A straightforward solution is for the clients to use MPC to compute the model's prediction from their embeddings on the model head. However, this solution, while more efficient than evaluating the complete model with MPC, still scales poorly. Even when the local client models have all the attention layers, the model head can still be prohibitively expensive to run, especially for larger models. 

To remedy this, we propose a new hybrid method, \textbf{Public/Private Hybrid Head-VFL (PPHH-VFL)}, in Section \ref{sec:methodology}, to efficiently evaluate the model head while preserving the empirical privacy of client embeddings. Our approach splits the model head into a large public head that runs in plaintext and a small private head that runs under MPC. The embeddings received by each head are also split into public and private components. As shown in previous work \cite{sun2022label, friedrich2019adversarial, jia2018attriguard, song2019overlearning, zhan2022privacy, huang2018generative}, we can limit privacy attacks on public embeddings through adversarial training. Adversarial training against privacy attacks reduces privacy leakage from public embeddings, thereby limiting their utility for the public head. 
The small private head compensates for this loss in utility by providing an avenue for sensitive information to flow into and influence the model's predictions, with minimal runtime overhead due to its small size. Hence, our approach combines empirical and theoretical privacy into a more efficient hybrid approach.

To empirically compare our approach, we define two baselines: an End-to-End MPC (E2E) baseline that infers a central model using only MPC, and a VFL+MPC baseline that combines VFL and MPC, as in our approach, but with a private head run exclusively under MPC. 
We experiment with two time-series datasets and models with parameter counts ranging from $1$ to $86$ million parameters and measure the runtime speedup provided by our approach over our baselines in both Local Area Networks (LAN) and Wide Area Networks (WAN). We find that our method improves runtime over the End-to-End MPC approach by up to \textbf{six orders of magnitude} $({1{,}570{,}000\times}$), turning a $\mathbf{157}$ \textbf{day} inference process into an $\mathbf{8.6}$ second one at the cost of a $10.21\%$ downstream classification accuracy drop. Against the VFL+MPC baseline, PPHH-VFL is competitive on small models, matching VFL+MPC. On larger models, PPHH-VFL's better scaling yields a runtime improvement of up to $\mathbf{44.4\times}$ in the WAN and $\mathbf{23.5\times}$ in the LAN, while improving downstream classification accuracy by $2.50\%$ and downstream regression RMSE by $40.7\%$. Finally, PPHH-VFL lowers communication cost for all model sizes over VFL+MPC by up to $\mathbf{91.2\times}$, with the communication cost per 64-sample batch for the largest model lowering from $\mathbf{1.7}$ \textbf{GB} to $\mathbf{19}$ \textbf{MB}.

We summarize our contributions as follows:
\begin{itemize}
    \item We present a new interpretation of the time-series VFL inference model, where time-steps rather than time-step features are split across clients.
    \item We propose a new method, \textbf{PPHH-VFL}, that splits the model head into a privacy-protecting, efficient public head and a small secure private head evaluated under MPC.
    \item We empirically compare against two baselines: End-to-End MPC and VFL+MPC, and show that PPHH-VFL enables practical runtimes and a significant reduction in communication cost. All the while improving downstream task performance relative to the closest-performing realistic baseline (VFL+MPC).
\end{itemize}

\section{Background}

\subsection{Vertical Federated Learning} \label{subec:federated_learning_background}
Federated Learning (FL) \cite{mcmahan2017communication} is a distributed machine learning approach where multiple clients collaboratively train a global model without centralizing their local datasets. Let $ w \in \mathbb{R}^k $ represent the set of weights for a model.

In Vertical Federated Learning (VFL) \cite{yang2019federated}, data record features are split between the clients. For example, if an individual's financial data is a data record, one bank might have credit card-related features, while another might have mortgage-related features for the same individual. To accommodate this distribution of features across clients, each data record is assigned a unique ID, and before inference of a VFL model, clients must run a privacy-preserving entity alignment protocol to ensure their IDs match. This step is usually performed using private set intersection techniques \cite{liu2024vertical}. Then, each client encodes its input features using their respective local model $w^i$ into a local embedding vector $e \in \mathbb{R}^{h}$, with $h$ the dimensionality of the embedding vector. Finally, the clients send their embedding to the server, which can either be a third-party or one of the clients themselves, which concatenates the embeddings and infers the model's prediction using the model head $w^S$. 

VFL requires collaboration for both training and inference, since the features for any data record remain split between clients at inference time. Our work focuses on inference specifically, so we detail below the inference process.
\begin{enumerate}
    \item \textbf{(Privacy-Preserving Entity Alignment)} The clients agree on the data record IDs to infer on and their order using private set intersection.
    \item \textbf{(Client Forward Phase)} Each client computes their local embedding $e_i$ with their local model $w^i_t$ and send it to the server.
    \item \textbf{(Server Forward Phase)} The server concatenates the client embeddings $e=[e_0, e_1, \dots,\\ e_{n-1}]$, computes the model's head prediction and sends the prediction back to the clients.
\end{enumerate}

\subsection{Multi-Party Computation}
A common technique to jointly compute a function while keeping the inputs to the function private is multi-party computation (MPC). Parties can either provide their input to the protocol in the form of secret shares (non-computational parties) or perform the computation using those shares (computational parties). In this work, we focus on the two-computational-party setting.

\textbf{Secret Sharing:} Privately share input data between $n$ parties such that all $n$ parties are required to decode it, meaning that even $n-1$ colluding malicious parties cannot reveal anything about a benign party's input. 

 \textbf{Arithmetic MPC \& Non-Linear Operations:} Arithmetic MPC protocols utilize a linear secret sharing scheme, such as an additive secret sharing scheme, to compute complex circuits via combinations of additions and multiplications. However, because they are limited to these basic operations, computing non-linear operations requires alternative techniques. For comparisons, converting to binary secret shares or using Yao's garbled circuits is common \cite{demmlerABYFrameworkEfficient2015}. To compute exponentiation and division, which are necessary for common ML operations like Softmax, popular MPC frameworks like CrypTen \cite{knottCrypTenSecureMultiParty2021} employ iterative arithmetic approximations. These approximations are highly communication-intensive, creating the bottleneck our work indirectly addresses.

\section{Problem Definition}\label{sec:prob_def}

We consider the setting where time-series sequences $S_0, S_1, \dots$ are shared between $n$ clients, where $S_i = [s_0^i, s_1^i, \dots, s_k^i]$, and $s^i_j \in \mathbb{R}^l$ is a set of features for the $j$-th time-step from sequence $i$. The clients want to infer a property from this sequence using a transformer model $M_\theta(S_i)$, for example, whether a transaction in the sequence is fraudulent.
We specifically study the setting in which each sequence $S_i = [s_0^i, s_1^i, \dots, s_k^i]$ is randomly partitioned between clients. For example, given a sequence $S_e = [s_0, s_1, s_2, s_3, s_4]$, client A owns time-steps $S_a = [s_0, s_2, s_3]$ and client B owns time-steps $S_b = [s_1, s_4]$ and together the want to compute $M_\theta(S_e)$. Our goal is to find an efficient method to minimize online inference runtime while preserving the privacy of the input sequence. We are willing to incur a small accuracy degradation to achieve usable runtimes. We are also willing to forego theoretical privacy guarantees to achieve these runtimes and limit the resulting degradation in accuracy.
We assume the clients have access to training data to train $M_\theta$. We use CrypTen as our MPC framework \cite{knottCrypTenSecureMultiParty2021}.


\section{Methodology}\label{sec:methodology}

We simplify the inference scenario by assuming that clients can, using MPC, privately recover the position of the time-steps they own in the complete sequence $S$, which is efficiently feasible with existing MPC frameworks like CrypTen \cite{knottCrypTenSecureMultiParty2021}. For example, given a sequence $S = [s_0, s_1, s_2, s_3, s_4]$, where client A owns time-steps $S_a = [s_0, s_2, s_3]$ and client B owns time-steps $S_b = [s_1, s_4]$, the clients have access to an MPC protocol $p$ that, given shares of $S_a$ and shares of $S_b$, returns the position of the each time step $s_i$ within $S$ for their respective sequence. In our example, using $p$, client A would learn that their time-steps are located at positions $[0,2,3]$ and client B would learn that their time-steps are located at positions $[1,4]$. While learning this information leaks some private information about other clients' data, this leakage is small and would be inherent in a system operating on live streaming data.
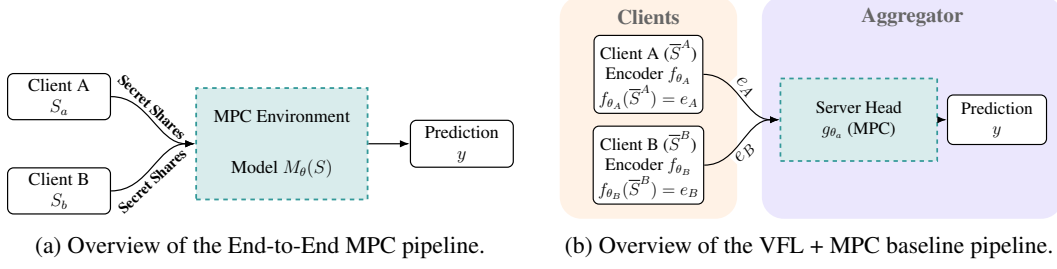
\begin{figure}[t] 
    \centering
    
    \begin{subfigure}[b]{0.48\textwidth}
        \centering
        \begin{adjustbox}{width=\columnwidth}
        \begin{tikzpicture}[
            box/.style={draw, rectangle, rounded corners, thick, fill=white, align=center, minimum height=0.8cm, minimum width=2.4cm, font=\large},
            mpc/.style={draw=teal!80, rectangle, dashed, very thick, fill=teal!15, align=center, minimum height=2.6cm, minimum width=4cm, font=\large},
            arrow/.style={-Latex, thick},
            labelstyle/.style={font=\normalsize\bfseries, inner sep=2pt, sloped}
        ]

        \node[box] (clientA) at (0, 1.1) {Client A \\ $S_a$};
        \node[box] (clientB) at (0, -1.1) {Client B \\ $S_b$};

        \node[mpc] (mpc_env) at (5.2, 0) {MPC Environment \\ \vspace{0.2cm} \\ Model $M_\theta(S)$};
        
        \node[box] (output) at (9.4, 0) {Prediction \\ $y$};

        \draw[arrow] (clientA.east) to[out=0, in=180] node[labelstyle, above, pos=0.4] {Secret Shares} (mpc_env.west);
        \draw[arrow] (clientB.east) to[out=0, in=180] node[labelstyle, below, pos=0.4] {Secret Shares} (mpc_env.west);
        
        \draw[arrow] (mpc_env.east) to[out=0, in=180] (output.west);

        \end{tikzpicture}
        \end{adjustbox}
        \caption{Overview of the End-to-End MPC pipeline.}
        \label{fig:end_to_end_mpc}
    \end{subfigure}
    \hfill
    \begin{subfigure}[b]{0.48\textwidth}
        \centering
        \begin{adjustbox}{width=\columnwidth}
        
        \definecolor{cbpurple}{HTML}{785EF0}
        
        \begin{tikzpicture}[
            box/.style={draw, rectangle, rounded corners, thick, fill=white, align=center, minimum height=1.2cm, minimum width=2.4cm, font=\large},
            mpc/.style={draw=teal!80, rectangle, dashed, very thick, fill=teal!15, align=center, minimum height=2.0cm, minimum width=3.8cm, font=\large},
            arrow/.style={-Latex, thick},
            labelstyle/.style={font=\Large\bfseries, fill=white, inner sep=2pt, sloped}
        ]

        \fill[orange!10, rounded corners=15pt] (-0.4, 3) rectangle (3.9, -2.4);
        \fill[cbpurple!15, rounded corners=15pt] (4.5, 3) rectangle (11.8, -2.4);
        
        \node[font=\Large\bfseries, text=gray!80!black] at (1.75, 2.5) {Clients};
        \node[font=\Large\bfseries, text=gray!80!black] at (8.15, 2.5) {Aggregator};

        \node[box] (clientA) at (1.75, 1.1) {Client A ($\overline{S}^A$)\\ Encoder $f_{\theta_A}$ \\ $f_{\theta_A}(\overline{S}^A) = e_A$};
        \node[box] (clientB) at (1.75, -1.1) {Client B ($\overline{S}^B$) \\ Encoder $f_{\theta_B}$ \\ $f_{\theta_B}(\overline{S}^B) = e_B$};

        \node[mpc] (serverHead) at (6.85, 0) {Server Head \\ $g_{\theta_a}$ (MPC)};
        \node[box] (output) at (10.2, 0) {Prediction \\ $y$};

        \draw[arrow] (clientA.east) to[out=0, in=180] node[labelstyle, above, pos=0.44] {$e_A$} (serverHead.west);
        \draw[arrow] (clientB.east) to[out=0, in=180] node[labelstyle, below, pos=0.44] {$e_B$} (serverHead.west);
        
        \draw[arrow] (serverHead.east) -- (output.west);

        \end{tikzpicture}
        \end{adjustbox}
        \caption{Overview of the VFL + MPC baseline pipeline.}
        \label{fig:vfl_mpc}
    \end{subfigure}
    
    \caption{Comparison of traditional end-to-end MPC (left) versus the VFL with secure head evaluation (right).}
    \label{fig:combined_mpc_comparison}
\end{figure}
\paragraph{Baseline 1: End-to-End MPC.}
Following previous work \cite{li2022mpcformer, wang2022characterization, ding2023east}, the first approach we consider is to infer $M_\theta$ on the sequence $S$ split between $n$ clients using MPC for the entire inference. Figure \ref{fig:end_to_end_mpc} shows an overview of the pipeline. Each client secret-shares its part of the input sequence $S_{c_i}$ with the other clients, and we assume the model is already secret-shared among them. The clients can then privately compute $M_\theta(S)$ using both our aforementioned sequence-ordering reconstruction protocol and standard ML inference MPC protocols \cite{li2022mpcformer, wang2022characterization, ding2023east}. 
However, this approach is prohibitively slow and infeasible in realistic settings, as corroborated by our results in Section \ref{sec:experiments}.

\paragraph{Baseline 2: VFL + MPC.}
To achieve usable runtimes, a straightforward modification of Baseline 1 proposed in previous work is to replace $M_\theta$ with a VFL model \cite{wu2023falcon} and use MPC only to compute the model head over the client's embeddings. This approach boasts minimal accuracy degradation and preserves most of the theoretical privacy guarantees of Approach 1. In more detail: a VFL model, $M_\theta$, is split into $n$ encoder models $f_{\theta_0}, f_{\theta_1}, \dots, f_{\theta_{n-1}}$, one for each client, and one model head for the server $g_{\theta_a}$. After using protocol $p$ to recover their time-steps' positions, each client $i$ can craft a masked version $\bar{S}^i$ of the full sequence $S$ where the other clients' entries are replaced with a MASK token value. 

Then, each client encodes its masked sequence $\bar{S}^i$ into an embedding $f_{\theta_i}(\bar{S}^i) = e_i$ locally, in plaintext. Finally, using MPC, the clients and the server can compute the final prediction on the model head: $g_{\theta_a}(e_0, e_1, \dots, e_{n-1})$. 
Figure \ref{fig:vfl_mpc} illustrates an overview of the process.
This approach leverages VFL to offer a significant speedup over Approach 1. Still, its runtime scales poorly with the number of clients and the embedding size, as these linearly increase the model head size evaluated under MPC, thereby disproportionately increasing runtime. Approach 2 also greatly limits model architecture design choices, as the model head needs to be a small Multi-Layer Perceptron (MLP) to achieve realistic runtimes. Having any transformer encoder/decoder layer in the model head would significantly slow down runtime \cite{li2022mpcformer, wang2022characterization, ding2023east}. Likewise, this approach limits the scale of solvable tasks, as more complex tasks often require larger models and, therefore, embeddings and model heads, which significantly increase runtime under MPC.

\paragraph{Our Approach: Public/Private Hybrid Head-VFL (PPHH-VFL).}
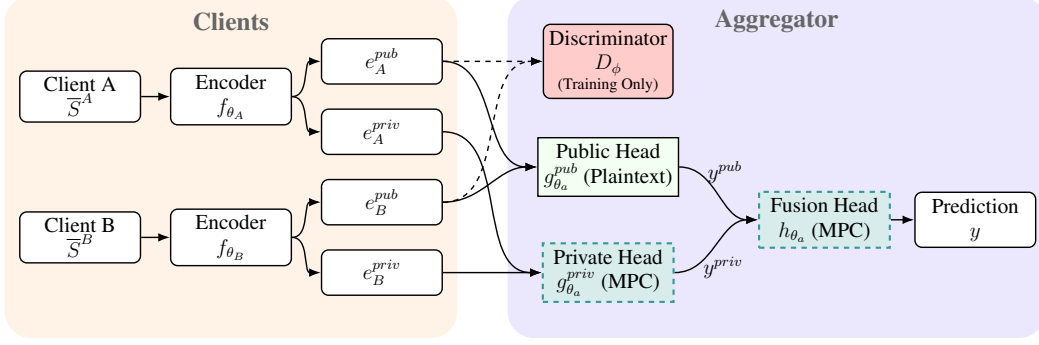
\begin{figure*}[t]
    \centering
    \begin{adjustbox}{width=0.99\textwidth}
    
    \definecolor{cbpurple}{HTML}{785EF0}
    
    \begin{tikzpicture}[
        box/.style={draw, rectangle, rounded corners, thick, fill=white, align=center, minimum height=0.9cm, minimum width=2.4cm, font=\large},
        mpc/.style={draw=teal!80, rectangle, dashed, very thick, fill=teal!15, align=center, minimum height=1cm, minimum width=2.6cm, font=\large},
        plaintext/.style={draw, rectangle, thick, fill=green!5, align=center, minimum height=1cm, minimum width=2.6cm, font=\large},
        arrow/.style={-Latex, thick},
        labelstyle/.style={font=\large\bfseries, inner sep=2pt}
    ]

    \fill[orange!10, rounded corners=15pt] (-1.5, 3.4) rectangle (7.5, -3.4);
    \fill[cbpurple!15, rounded corners=15pt] (8.5, 3.4) rectangle (19.2, -3.4);
    
    \node[font=\Large\bfseries, text=gray!80!black] at (3, 2.9) {Clients};
    \node[font=\Large\bfseries, text=gray!80!black] at (13.85, 2.9) {Aggregator};

    \node[box] (clientA) at (0, 1.4) {Client A\\ $\overline{S}^A$};
    \node[box] (clientB) at (0, -1.4) {Client B \\ $\overline{S}^B$};

    \node[box] (encA) at (3, 1.4) {Encoder \\ $f_{\theta_A}$};
    \node[box] (encB) at (3, -1.4) {Encoder \\ $f_{\theta_B}$};

    \node[box] (pubA) at (6, 2.1) {$e_A^{pub}$};
    \node[box] (privA) at (6, 0.7) {$e_A^{priv}$};
    \node[box] (pubB) at (6, -0.7) {$e_B^{pub}$};
    \node[box] (privB) at (6, -2.1) {$e_B^{priv}$};

    \node[box, fill=red!20] (disc) at (10.5, 2.1) {Discriminator \\ $D_\phi$ \\[-0.1cm] \small (Training Only)};
    \node[plaintext] (pubHead) at (10.5, 0.0) {Public Head \\ $g_{\theta_a}^{pub}$ (Plaintext)};
    \node[mpc] (privHead) at (10.5, -2.1) {Private Head \\ $g_{\theta_a}^{priv}$ (MPC)};

    \node[mpc] (fusionHead) at (14.8, -1.05) {Fusion Head \\ $h_{\theta_a}$ (MPC)};

    \node[box] (output) at (17.8, -1.05) {Prediction \\ $y$};

    \draw[arrow] (clientA.east) -- (encA.west);
    \draw[arrow] (clientB.east) -- (encB.west);

    \draw[arrow] (encA.east) to[out=0, in=180] (pubA.west);
    \draw[arrow] (encA.east) to[out=0, in=180] (privA.west);
    \draw[arrow] (encB.east) to[out=0, in=180] (pubB.west);
    \draw[arrow] (encB.east) to[out=0, in=180] (privB.west);

    \draw[arrow, dashed] (pubA.east) to[out=0, in=180] (disc.west);
    \draw[arrow, dashed] (pubB.east) to[out=0, in=180] (disc.west);

    \draw[arrow] (pubA.east) to[out=0, in=180] (pubHead.west);
    \draw[arrow] (pubB.east) to[out=0, in=180] (pubHead.west);
    \draw[arrow] (privA.east) to[out=0, in=180] (privHead.west);
    \draw[arrow] (privB.east) to[out=0, in=180] (privHead.west);

    \draw[arrow] (pubHead.east) to[out=0, in=180] node[labelstyle, above, yshift=8pt, pos=0.6] {$y^{pub}$} (fusionHead.west);
    \draw[arrow] (privHead.east) to[out=0, in=180] node[labelstyle, below, yshift=-8pt, pos=0.6] {$y^{priv}$} (fusionHead.west);
    
    \draw[arrow] (fusionHead.east) -- (output.west);

    \end{tikzpicture}
    \end{adjustbox}
    \caption{Overview of the PPHH-VFL pipeline utilizing the public/private head split and MPC fusion.}
    \label{fig:pphh_vfl}
\end{figure*}
Our approach builds on Approach 2, with one crucial architectural change: we split the model head held by the server into a small private head, $g^{priv}_{\theta_a}$, a large public head, $g^{pub}_{\theta_a}$, and a tiny fusion head $h_{\theta_a}$. To accommodate this change, each client now separates their embeddings into two components as well: $f_{\theta_i}(\bar{S}^i) = (e^{priv}_i, e^{pub}_i)$. Then, the server computes the private head's prediction on the private embeddings using MPC: $g_{\theta_a}^{priv}(e^{priv}_0, e^{priv}_1, \dots, e^{priv}_{n-1}) = y^{priv}$ and the public head's prediction on the public embeddings in plaintext: $g_{\theta_a}^{pub}(e^{pub}_0, e^{pub}_1, \dots, e^{pub}_{n-1}) = y^{pub}$. These two computations can occur in parallel. 
Finally, the model prediction is computed by passing the public and private predictions through a model fusion layer using MPC:  $h_{\theta_a}(y^{pub}, y^{priv}) = y$.
Figure \ref{fig:pphh_vfl} shows an overview of the pipeline.

Reducing the size of the model evaluated under MPC, without changing the overall size of the model head, provides a significant speedup over Approach 2 without degrading accuracy. However, privacy leakage from the public head evaluated in plaintext could be used by clients to mount privacy attacks against each other \cite{morris2023text, sun2022label, li2023sentence}. To prevent this, we follow previous work \cite{sun2022label, friedrich2019adversarial, jia2018attriguard, song2019overlearning, zhan2022privacy, huang2018generative} and adversarially train the client's encoder models to limit leakage from public embeddings. 

To provide meaningful privacy via adversarial training, we use a machine learning model, called the discriminator, to model arbitrary privacy attacks. The discriminator's sole purpose is to determine the originator among the clients of a public embedding, given that embedding. The intuition behind this privacy design is that if one cannot distinguish which institution sent the embedding, one cannot infer information about the client's data that is used solely by that institution to create the embedding, since it is only necessary to detect the presence of such information to distinguish the sources. We train the discriminator in tandem with our VFL model, similarly to a Generative Adversarial Network (GAN) \cite{goodfellow2020generative}. If, by the end of training, the discriminator cannot achieve better than random predictions, we deem our approach empirically private. 

We use the cross-entropy loss \cite{bridle1990probabilistic} to train the discriminator $D_{\phi}$:
\begin{equation}
\mathcal{L}_{D} = \frac{1}{n}\sum_{i=0}^{n-1}CE(D_{\phi}(e^{pub}_i), i)    
\end{equation}
When training our VFL model $M_\theta$, we add a privacy-preserving loss component $\mathcal{L}_{priv}$ to the underlying task loss $\mathcal{L}_{task}$ to obtain the model loss $\mathcal{L}_{G}$. For fraud-detection, we use the binary cross-entropy loss for the task loss $\mathcal{L}_{task}$.
\begin{equation}
\mathcal{L}_{G} = \mathcal{L}_{task} + \alpha \mathcal{L}_{priv}
\end{equation}
Where $\alpha$ is a scalar affecting the strength of the privacy loss component.
We compute the privacy-preserving loss component as follows:
\begin{equation}
    \mathcal{L}_{priv} = -\mathcal{L}_D  
\end{equation}

The discriminator and generator have opposing losses, ensuring that one of them must succeed. We expect that the task of generating embeddings with minimal privacy leakage that are useful for inference, while having access to a private channel where information can flow freely, is a setting that advantages the generator over the discriminator. 
We empirically validate this claim in Section \ref{subsec:emp_priv}. Depending on the specific privacy concerns, one could adapt the discriminator's loss $\mathcal{L}_{D}$ and the privacy-preserving loss component $\mathcal{L}_{priv}$ accordingly to achieve the desired empirical privacy.

If our public head has limited privacy leakage, one might wonder about the use of the model's private head. While it can be the case that adversarial training maintains accuracy for simpler tasks \cite{liu2019privacy, zhan2022privacy}, adversarial training can decrease performance for more complex tasks \cite{tsipras2018robustness}. Therefore, having a small avenue for unconstrained, private information to flow through the private head and affect the final prediction improves task performance with minimal overhead at inference time.

\section{Experiments} \label{sec:experiments}
In this section, we provide an end-to-end comparison of our PPHH-VFL approach against the two baselines introduced in Section~\ref{sec:methodology}: End-to-End MPC and VFL+MPC. We evaluate inference speed, communication cost, and downstream utility under different network conditions. We begin with the experimental setup, then compare the three approaches in terms of runtime and downstream utility. Section \ref{subsec:perf_eval} showcases the advantage of PPHH-VFL over the baselines in terms of runtime and communication cost across two datasets, especially for larger model sizes. We then present the downstream utility of each method in Section \ref{subsec:downstream_utility}, showing that PPHH-VFL offers competitive utility while offering a significant speedup over other approaches. Finally, we analyze the performance of our generator-discriminator training and demonstrate the empirical privacy of our approach. We provide further experimentation in Appendix \ref{app:add_experiments}.

\subsection{Experimental Setup}\label{subsubsec:exp_setup}

In our experiments, we vary model and embedding sizes to measure their effect on runtime and communication cost. For VFL+MPC, we ascribe $h_{priv}$ as the size of the embedding generated by each client, and $a_{priv}$ as the size of the hidden layers of the model head evaluated under MPC. For PPHH-VFL, $h_{pub}$ and $h_{priv}$ represent the public and private embedding sizes, respectively, while $a_{pub}$ and $a_{priv}$ represent the size of the hidden layers of the public and private head, respectively. We design four architecture for VFL+MPC (P1, P2, P3, P4) and PPHH-VFL (H1, H2, H3, H4) such that the layer structure and the number of parameters matches as closely as possible between equivalent architectures (P1 $\simeq$ H1, P2 $\simeq$ H2, \dots). Our architectures range from 1 to 86 million trainable parameters.  We consider two networking profiles: LAN with 1 ms Round Trip Time (RTT) and 1 Gbps bandwidth, and WAN with 40 ms RTT and 100 Mbps bandwidth.
We provide additional implementation details and hyperparameters in Appendix \ref{app:setup_info}.

\subsection{Performance Evaluation}\label{subsec:perf_eval}
We use the Credit Card Transactions Fraud Detection Dataset on \href{https://www.kaggle.com/datasets/kartik2112/fraud-detection}{Kaggle} using the provided training and test splits.
The dataset is synthetic and was generated using the \href{https://github.com/namebrandon/Sparkov_Data_Generation}{Sparkov} data generation tool. The dataset contains 1000 transaction sequences, one for each customer, across 800 merchants. The classification task we train on is binary: whether a client committed a fraudulent transaction. The maximum sequence length for this dataset is 1534 time-steps.
To further validate our empirical results, we evaluate our approach on the Rossmann Sales Dataset \cite{rossmann}, which is also publicly available on \href{https://www.kaggle.com/competitions/rossmann-store-sales}{Kaggle}.
The dataset consists of real Rossmann store data published as a forecast competition by the company, with the goal of predicting 6 weeks of daily sales for 1115 stores across Germany. The regression task we train on is to predict the last day of sales given all previous days. The maximum sequence length for the Rossmann dataset is 942 time steps.

\begin{table}[t]
\centering
\small
\setlength{\tabcolsep}{3pt}
\caption{Online inference runtime across model sizes and network profiles for the Fraud and Rossmann datasets. We report \textit{s/batch} using 3 batches of size 1 for E2E MPC, 15 of size 64 for the rest of Fraud, and 2 of size 64 for Rossmann. We specify the architecture used, with the associated $(h_{\mathrm{priv}}, a_{\mathrm{priv}})$ for VFL+MPC, and $(h_{\mathrm{pub}}, a_{\mathrm{pub}} \mid h_{\mathrm{priv}}, a_{\mathrm{priv}})$ for PPHH-VFL. The results are averaged over 9 runs and we include the 95\% confidence intervals. We bold the best performer between equivalent model architectures (P1 $\simeq$ H1, P2 $\simeq$ H2, \dots).}
\label{tab:runtime_table}
\begin{tabular}{llrr}
\toprule
Model & Architecture & LAN s/batch & WAN s/batch \\
\midrule
\textbf{Fraud: E2E MPC} &  & 13880.06 $\pm$ 2.84 & DNF \\
\multicolumn{4}{l}{\textbf{Fraud: VFL+MPC}} \\
P1 & $(64, 64)$ & \textbf{0.45 $\pm$ 0.01} & 3.17 $\pm$ 0.03 \\
P2 & $(256, 256)$ & 0.54 $\pm$ 0.02 & 3.13 $\pm$ 0.01 \\
P3 & $(1024, 1024)$ & 1.84 $\pm$ 0.02 & 14.86 $\pm$ 0.01 \\
P4 & $(4096, 4096)$ & 15.05 $\pm$ 0.02 & 139.44 $\pm$ 0.01 \\
\addlinespace
\multicolumn{4}{l}{\textbf{Fraud: PPHH-VFL}} \\
H1 & $(48, 64 \mid 16, 32)$ & 0.54 $\pm$ 0.01 & \textbf{3.10 $\pm$ 0.02} \\
H2 & $(192, 256 \mid 32, 64)$ & 0.54 $\pm$ 0.01 & 3.13 $\pm$ 0.02 \\
H3 & $(768, 1024 \mid 64, 128)$ & \textbf{0.55 $\pm$ 0.00} & \textbf{3.09 $\pm$ 0.02} \\
H4 & $(3072, 4096 \mid 128, 256)$ & \textbf{0.64 $\pm$ 0.01} & \textbf{3.14 $\pm$ 0.02} \\
\addlinespace
\multicolumn{4}{l}{\textbf{Rossmann: VFL+MPC}} \\
P1 & $(64, 64)$ & \textbf{0.70 $\pm$ 0.11} & 4.33 $\pm$ 0.27 \\
P2 & $(256, 256)$ & \textbf{0.95 $\pm$ 0.11} & \textbf{3.99 $\pm$ 0.10} \\
P3 & $(1024, 1024)$ & 2.43 $\pm$ 0.15 & 15.82 $\pm$ 0.07 \\
P4 & $(4096, 4096)$ & 17.94 $\pm$ 0.11 & 142.80 $\pm$ 0.06 \\
\addlinespace
\multicolumn{4}{l}{\textbf{Rossmann: PPHH-VFL}} \\
H1 & $(48, 64 \mid 16, 32)$ & 1.95 $\pm$ 0.39 & \textbf{3.85 $\pm$ 0.04} \\
H2 & $(192, 256 \mid 32, 64)$ & 1.68 $\pm$ 0.02 & 4.07 $\pm$ 0.11 \\
H3 & $(768, 1024 \mid 64, 128)$ & \textbf{1.70 $\pm$ 0.04} & \textbf{4.09 $\pm$ 0.08} \\
H4 & $(3072, 4096 \mid 128, 256)$ & \textbf{2.11 $\pm$ 0.05} & \textbf{4.47 $\pm$ 0.06} \\
\bottomrule
\end{tabular}
\end{table}
\paragraph{Inference Time - Fraud Dataset.} Table~\ref{tab:runtime_table} reports the measured online runtime for each configuration and model architecture with three clients. First, we find that End-to-End MPC, which we run with the smallest architecture, is entirely impractical for two reasons: it requires up to $\mathbf{1{,}570{,}000\times}$ the runtime of PPHH-VFL for the entire Fraud dataset, and running the transformer encoder layers under MPC has a large memory footprint, allowing only a batch size of 1 even with 80GB of GPU memory. Inference on the entire Fraud dataset would take over \textbf{150 days} in LAN for End-to-End MPC on the smallest model architecture, while for the largest model, VFL+MPC would take $\mathbf{\sim 4}$ \textbf{minutes} and PPHH-VFL would take $\mathbf{\sim 10}$ \textbf{seconds}. Therefore, for the remainder of our experiments, we do not consider End-to-End MPC. \\
Second, for small models, namely P(1-2) and H(1-2), we find that VFL+MPC and PPHH-VFL perform similarly, except that VFL+MPC is $14.8\%$ faster in LAN for P1/H1 and PPHH-VFL is $2.2\%$ faster in WAN for P1/H1. This finding is expected for small models, as the computation overhead of MPC is minimal. \\
Finally, as we increase the model size, we find that, for VFL+MPC, the runtime per batch increases by up to $33.4 \times$ in the LAN and $44.5 \times$ in the WAN. For PPHH-VFL, the runtime per batch increases by up to $1.2\times$ in the LAN and by less than $2\%$ in the WAN. These results demonstrate that our approach, PPHH-VFL, scales significantly better to larger model sizes than VFL+MPC. For the largest model ($\sim$ 86M parameters), PPHH-VFL runs $\mathbf{23.5\times}$ faster than VFL+MPC in the LAN, and $\mathbf{44.4\times}$ faster than VFL+MPC in the WAN. These results demonstrate PPHH-VFL's runtime advantage for larger model sizes, which is especially relevant with the dominance of larger transformer models in recent research.

\paragraph{Inference Time - Rossmann Dataset.} For the Rossmann dataset, also reported in Table~\ref{tab:runtime_table}, we find that runtimes behave similarly to those of the Fraud Dataset, with VFL+MPC scaling poorly as model size increases, while PPHH-VFL remains nearly as fast. For small models, VFL+MPC is up to $2.8\times$ faster than PPHH-VFL in the LAN, while PPHH-VFL is up to $1.12\times$ faster in the WAN. However, for the largest models, PPHH-VFL achieves a speedup of up to $9.8\times$ in the LAN and $42.8\times$ in the WAN.

\begin{table*}[t]
\centering
\small
\caption{Communication cost per batch under LAN and WAN for a batch of size 64 on the Fraud dataset. We bold the best performer between equivalent model architectures (P1 $\simeq$ H1, P2 $\simeq$ H2, \dots).}
\label{tab:microbench_comm}
\begin{tabular}{lcccccccc}
\toprule
Stage & P1 & P2 & P3 & P4 & H1 & H2 & H3 & H4 \\
\midrule
Comm rounds & \textbf{32} & \textbf{32} & \textbf{32} & \textbf{32} & 33 & 33 & 33 & 33 \\
Comm (MB) & 6.42 & 29.63 & 181.42 & 1732.31 & \textbf{2.31} & \textbf{4.44} & \textbf{8.97} & \textbf{19.00} \\
Comm time LAN (s) & \textbf{0.09} & 0.17 & 1.38 & 13.84 & 0.10 & \textbf{0.10} & \textbf{0.11} & \textbf{0.15} \\
Comm time WAN (s) & \textbf{2.60} & \textbf{2.61} & 14.26 & 138.35 & 2.68 & 2.68 & \textbf{2.68} & \textbf{2.68} \\
\bottomrule
\end{tabular}
\end{table*}

\paragraph{Communication Cost - Fraud Dataset.}
Table~\ref{tab:microbench_comm} shows that the communication advantage of PPHH-VFL grows rapidly with model size. Although PPHH-VFL uses one additional communication round to send the plaintext embeddings, it transmits far less data because only the small private and fusion heads are evaluated under MPC. At the largest model size, PPHH-VFL only needs to communicate \textbf{19 MB} for a batch, reducing communication size by $\mathbf{91.2\times}$ compared to VFL+MPC, which needs to communicate \textbf{1.7 GB} for the same batch. Therefore, in settings where communication costs are important or bandwidth is limited, PPHH-VFL is a more practical solution. This reduction of communication size translates into a reduction in communication time of up to 92.3$\times$ in the LAN and up to 51.6$\times$ in the WAN.

\subsection{Downstream Utility}\label{subsec:downstream_utility}
We present the best accuracy on the Fraud dataset and Root Mean Square Error (RMSE) on the Rossmann Sales for each approach in Table \ref{tab:utility}. We find that End-to-End MPC, which utilizes a central transformer model, outperforms PPHH-VFL by $10.21\%$ and VFL+MPC by $12.71\%$ in accuracy on the Fraud dataset. We attribute this stark degradation in VFL model accuracy to the distributed attention mechanism in VFL compared to a central model. In End-to-End MPC, the transformer attention layer computes attention over the entire sequence. In contrast, in the VFL setting, attention is computed only over the subset of elements for each client. Since the model head lacks an attention layer, this limits the capabilities of VFL models. Moving some of the transformer layers from the local client models to the head model is a solution to improve accuracy and match the End-to-End MPC, which we leave for future work. We point out that doing so would significantly slow down the runtime of the VFL+MPC approach, as attention layers are notoriously slow under MPC \cite{li2022mpcformer, wang2022characterization}. However, PPHH-VFL's runtime would retain its speed, as we could place the attention layers exclusively in the public head of the model, avoiding them under MPC.

Notably in Table \ref{tab:utility}, we notice that PPHH-VFL outperforms VFL+MPC by $2.50\%$ accuracy on the Fraud dataset and achieves a RMSE that is $1134.63$ lower than VFL+MPC (a $40.7\%$ reduction). We posit that this difference in downstream utility is due to the adversarial training of our public embeddings, which acts as a form of regularization, forcing it to discard sensitive private information that is not useful for the downstream task, as seen in adversarial representation learning research \cite{altinisik2023impact,sadeghi2019global}. We provide additional utility results in Appendix \ref{app:add_utility}.

\subsection{Empirical Privacy Analysis} \label{subsec:emp_priv}
\begin{figure}[t]
    \centering
    \begin{minipage}[c]{0.48\linewidth}
        \centering
        \captionof{table}{Best predictive performance across evaluated model sizes for each approach.}
        \label{tab:utility}
        \small
        \setlength{\tabcolsep}{3pt} 
        \begin{tabularx}{\linewidth}{l >{\centering\arraybackslash}X >{\centering\arraybackslash}X}
        \toprule
        Method & Fraud Accuracy (\%) & Rossmann RMSE \\
        \midrule
        E2E MPC & 86.88 & -- \\
        VFL+MPC & 74.17 & 2786.87 \\
        PPHH-VFL & 76.67 & 1652.24 \\
        \bottomrule
        \end{tabularx}
    \end{minipage}
    \hfill 
    \begin{minipage}[c]{0.48\linewidth}
        \centering
        \includegraphics[width=\linewidth]{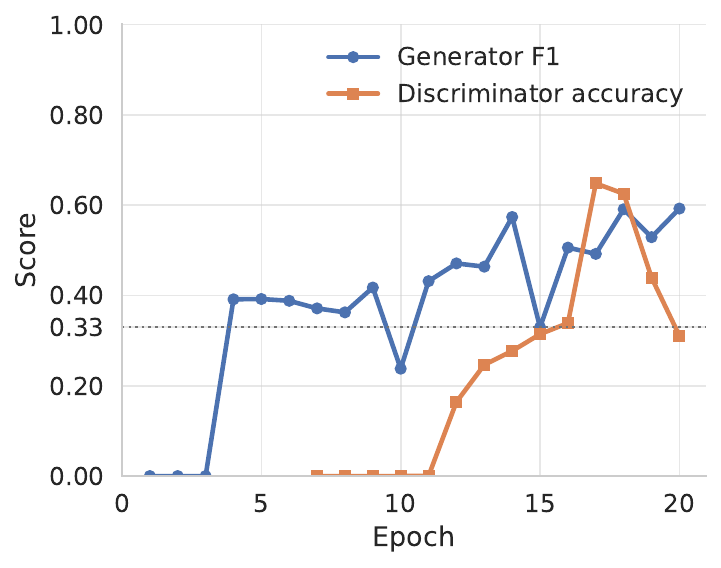}
        \captionof{figure}{Generator F1 score and discriminator accuracy during training, with the dotted line at 33.3\% representing random guessing for the discriminator.}
        \label{fig:epoch_vs_accuracy}
    \end{minipage}
    
\end{figure}

    

To measure the empirical privacy of public embeddings, we evaluate the discriminator's ability to identify the originating client of a public embedding. On the H1 model size, Figure \ref{fig:epoch_vs_accuracy} shows that the discriminator attains a per-class accuracy of $31.1\%$ compared to the random baseline of $33.3\%$. This result suggests that the public embeddings reveal no information about the source client's data. These results support the claim that adversarial training can significantly reduce privacy attacks from the public embeddings while retaining enough utility for accurate inference.

\section{Related Work}
Private inference in VFL is a well-studied problem for linear and logistic regression \cite{hardy2017private, liu2024vertical, shankar2025share, tang2023ihvfl} and tree-based models \cite{cheng2021secureboost, liu2024vertical, shankar2025share}, with many works using either Homomorphic Encryption (HE), MPC, or a combination of both \cite{wu2023falcon, liu2024vertical}. For neural network models, private inference in VFL is more challenging in terms of computation and communication, and related work has focused mainly on shallow networks \cite{liu2024vertical} or Recurrent Neural Network (RNN) architectures \cite{yan2022multi}. While there has been work to speed up MPC inference for transformer-based models \cite{li2022mpcformer, wang2022characterization} or to speed up the activation-function bottleneck in transformer models \cite{ding2023east}, runtimes remain prohibitive and out of reach for practical use cases with large models where latency is a concern. 

The scenario in which features within each time step are split between clients has recently received attention from the research community \cite{shankar2025share, yan2022multi}. However, our scenario in which the entire sequence is treated as a single input and the time steps are split across clients remains, to our knowledge, unstudied. 
A complementary line of research investigates mitigating privacy leakage from embeddings rather than improving MPC protocols. Adversarial training has been proposed as a mechanism for learning embeddings that preserve task-relevant information while suppressing sensitive attributes. 
A common empirical privacy approach is to use a discriminator model as a privacy adversary to enable adversarial training and limit privacy leakage from model outputs while maintaining downstream performance \cite{liu2019privacy, zhan2022privacy}
This approach has already been used to represent human mobility trajectories (time-series data) \cite{zhan2022privacy} in the centralized setting using LSTM-based models. We apply a similar adversarial training process to transformer-based models and embeddings.

An alternative to MPC and empirical privacy is statistical perturbation, such as applying Local Differential Privacy (LDP) to the client embeddings before sending them to the server. In LDP, each client adds noise to its embedding before sending it to the server, ensuring that private information is provably hidden. However, LDP is known to have a strong negative impact on model accuracy in deep learning \cite{arachchige2019local} and was originally designed for low-dimensional data \cite{cao2026ldp}: the magnitude of the required noise scales with the size of the embedding vector \cite{feyisetan2021private} and is intended to hide the vector itself, i.e, to be effective, the noised vector needs to be as close to uniformly random as possible. Therefore, in our setting, where we have high-dimensional embedding vectors and need to hide the vector itself, LDP would significantly degrade accuracy and is not a practical option. 

\section{Conclusion}
In this work, we propose \textbf{Public/Private Hybrid Head-VFL (PPHH-VFL)}. Our hybrid VFL architecture splits the model head into an efficient and empirically private public head and a lightweight MPC private head. We also present a new interpretation of the VFL setting for time-series data, where entire time-steps are distributed across clients rather than their features. 
We evaluate our approach and baselines on two time-series datasets: the Kaggle Credit Card Transactions Fraud Detection dataset and the Rossmann Sales dataset. For both datasets, PPHH-VFL enables efficient, scalable inference for large transformer models where end-to-end MPC would be infeasible. PPHH-VFL also offers significant reductions in runtime and communication costs compared to the standard VFL+MPC baseline, while improving downstream task performance. 
As research into adversarially robust and private embeddings evolves, PPHH-VFL will directly benefit. Further study of other architectural splits, such as incorporating attention layers into the public model head, could bridge the downstream utility gap with the end-to-end MPC approach while retaining PPHH-VFL's low computation and communication overhead. Reproducing our results for other data modalities, such as text, audio, or image transformer models, could help display the utility of PPHH-VFL.  

\begin{ack}
We gratefully acknowledge the support of NSERC for grants RGPIN-2023-03244, IRC-537591, the Government of Ontario and the Royal Bank of Canada for funding this research.
\end{ack}



{
\small

\bibliographystyle{unsrt}
\bibliography{main}

}


\appendix

\section{Additional Experimental Setup Information}\label{app:setup_info}
\noindent \textbf{Experiment Details.}\quad All experiments are run on the same machine with 32 CPU cores @ 3.7 GHz and 1 TB of RAM with NVIDIA A100 GPUs with 80 GB of memory. Each client process is assigned a dedicated GPU, and we use a two-party computation setting. To simulate realistic network conditions, we execute all processes inside a Linux network namespace and enforce latency and bandwidth constraints on the loopback interface using Linux traffic control (\texttt{tc}). Specifically, we use \texttt{netem} to introduce delay and \texttt{tbf} to limit bandwidth.

\noindent \textbf{Hyperparameters.}\quad We use CrypTen with its default precision and settings. We keep the transformer model fixed across all methods with model dimension 64, 4 attention heads, 4 encoder layers, and feedforward dimension 512. The End-to-End model is trained with Adam, learning rate $3\times10^{-4}$, weight decay $10^{-5}$, batch size 32, and 10 epochs. The VFL+MPC baseline is trained with AdamW, learning rate $10^{-4}$, weight decay $10^{-6}$, batch size 64, and 20 epochs, with a ReduceLROnPlateau scheduler. PPHH-VFL is trained jointly with its discriminator using AdamW with generator and discriminator learning rates of $2\times10^{-4}$ and $7\times10^{-5}$ respectively, corresponding weight decays of $3\times10^{-7}$ and $8\times10^{-5}$, batch size 64, and 20 epochs. We detail the hyperparameters and exact training setup in our source code repository, available in the supplementary material or at the following anonymized link: \url{https://anonymous.4open.science/r/Private-VFL-Inference-TS-FE27}.

\section{Additional Experiments}\label{app:add_experiments}
\begin{table*}[t]
    \centering
    \small
    \setlength{\tabcolsep}{6pt} 
    \caption{Microbenchmark of the \% of runtime for each inference stage for PPHH-VFL under LAN and WAN. Averaged over 3 runs.}
    \label{tab:microbench_hybrid_combined}
    \begin{tabular}{lcccccccc}
    \toprule
     & \multicolumn{4}{c}{LAN} & \multicolumn{4}{c}{WAN} \\
    \cmidrule(lr){2-5} \cmidrule(lr){6-9}
    Stage & H1 & H2 & H3 & H4 & H1 & H2 & H3 & H4 \\
    \midrule
    Total MPC inference          & 57.6 & 57.9 & 58.8 & 61.2  & 95.2 & 95.0 & 94.9 & 94.7 \\
    \quad Private head forward   & 35.3 & 35.7 & 36.7 & 42.2  & 60.5 & 60.9 & 60.3 & 60.2 \\
    \quad Fusion head forward    & 19.8 & 19.7 & 19.8 & 16.9  & 31.8 & 31.2 & 31.7 & 31.5 \\
    \quad Output reveal          & 1.0  & 1.0  & 1.0  & 0.8   & 2.8  & 2.7  & 2.8  & 2.8  \\
    \quad Reconstruct input      & 1.4  & 1.5  & 1.4  & 1.3   & 0.2  & 0.2  & 0.2  & 0.2  \\
    Local transformer forward    & 40.4 & 40.6 & 35.9 & 33.6  & 4.3  & 4.3  & 4.2  & 4.3  \\
    Communication/share handling & 1.3  & 1.0  & 5.0  & 4.8   & 0.4  & 0.5  & 1.0  & 1.3  \\
    Public head forward          & 0.7  & 0.4  & 1.5  & 1.5   & 0.1  & 0.2  & 0.2  & 0.2  \\
    \bottomrule
    \end{tabular}
\end{table*}
\subsection{Microbenchmarks - Fraud Dataset}
We microbenchmark the runtime of each inference stage for PPHH-VFL in the LAN and the WAN, and present the results in Table \ref{tab:microbench_hybrid_combined}. Our approach has so little MPC overhead that $\sim 40\%$ of the runtime is spent on the local transformer, and the public head requires almost no runtime in the LAN. In the LAN, the MPC's share of runtime increases only slightly as model size increases, indicating that PPHH-VFL enables smooth scaling of model size without creating an MPC bottleneck. Unsurprisingly, in the WAN setting, since we increase the RTT from 1ms to 40ms, significantly more of the runtime is consumed by the MPC inference, roughly $\sim 95\%$. Since PPHH-VFL uses only 19 MB of communication for the largest model, reduced WAN bandwidth does not affect its performance.

\subsection{Varying the number of non-computational parties}

\begin{figure}[t]
\centering
\begin{subfigure}[t]{0.49\textwidth}
    \centering
    \includegraphics[width=\linewidth]{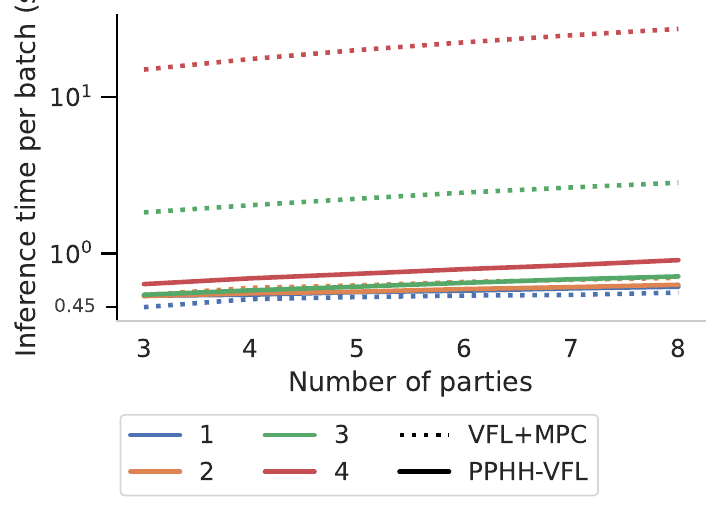}
    \caption{LAN}
    \label{fig:parties_vs_time_a}
\end{subfigure}
\hfill
\begin{subfigure}[t]{0.49\textwidth}
    \centering
    \includegraphics[width=\linewidth]{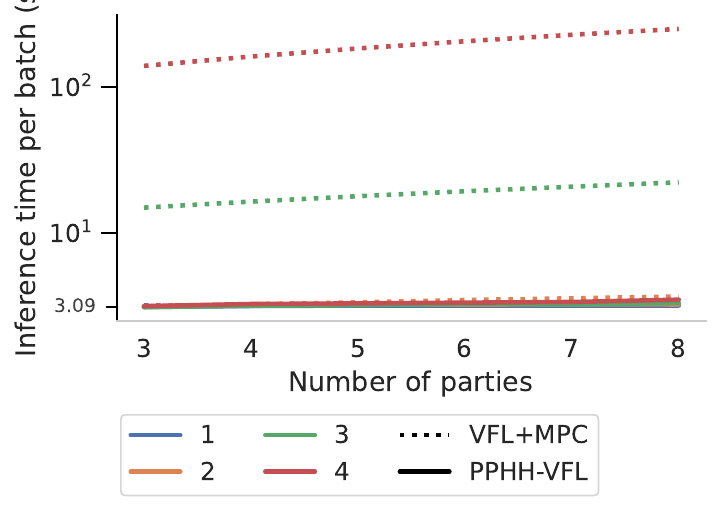}
    \caption{WAN}
    \label{fig:parties_vs_time_b}
\end{subfigure}
\caption{Online inference runtime in seconds (log-scale) as the number of parties scales for a batch size of 64 on the Fraud dataset for each model size.}
\label{fig:parties_vs_time}
\end{figure}

We want to measure how well the runtimes of PPHH-VFL and MPC+VFL scale with the number of clients (i.e., non-computational parties). We plot the inference time of PPHH-VFL and VFL+MPC across different parties and model sizes on the Fraud dataset in Figure \ref{fig:parties_vs_time}, with the number of parties ranging from three to eight. We find that, for the relative runtime increase from three to eight parties, the worst case for both approaches is the largest model size (P4 and H4) in both LAN and WAN. However, VFL+MPC scales more poorly than PPHH-VFL as the number of parties increases. For the largest model, VFL+MPC goes from $15.05$s to $27.41$s in the LAN, a $82.05\%$ increase; while PPHH-VFL goes from $0.64$s to $0.91$s, a $42.32\%$ relative increase, nearly half the relative increase of VFL+MPC.
The scaling gap widens in the WAN, with VFL+MPC going from $139.44$s to $249.98$s, a $79.27\%$ relative increase; while PPHH-VFL goes from $3.14$s to $3.47$s, a $10.76\%$ relative increase, which is $\mathbf{7.4}\times$ better than VFL+MPC.
Therefore, in addition to being faster, PPHH-VFL also offers better scaling than VFL+MPC as the number of non-computational parties increases. 

\subsection{Additional downstream utility and empirical privacy analysis results} \label{app:add_utility}
\begin{figure}[t]
    \centering
    \begin{subfigure}[b]{0.49\linewidth}
        \centering
        \includegraphics[width=\linewidth]{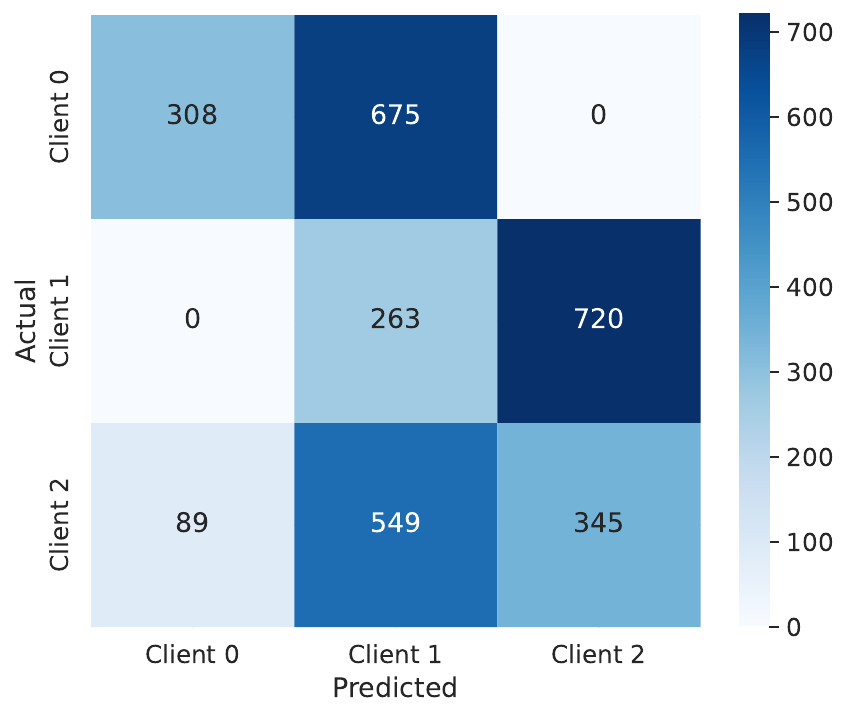}
        \caption{Confusion matrix of the H1 discriminator model predicting the client of origin for embeddings.}
        \label{fig:disc_confusion_matrix}
    \end{subfigure}
    \hfill 
    \begin{subfigure}[b]{0.49\linewidth}
    \centering
    \includegraphics[width=\linewidth]{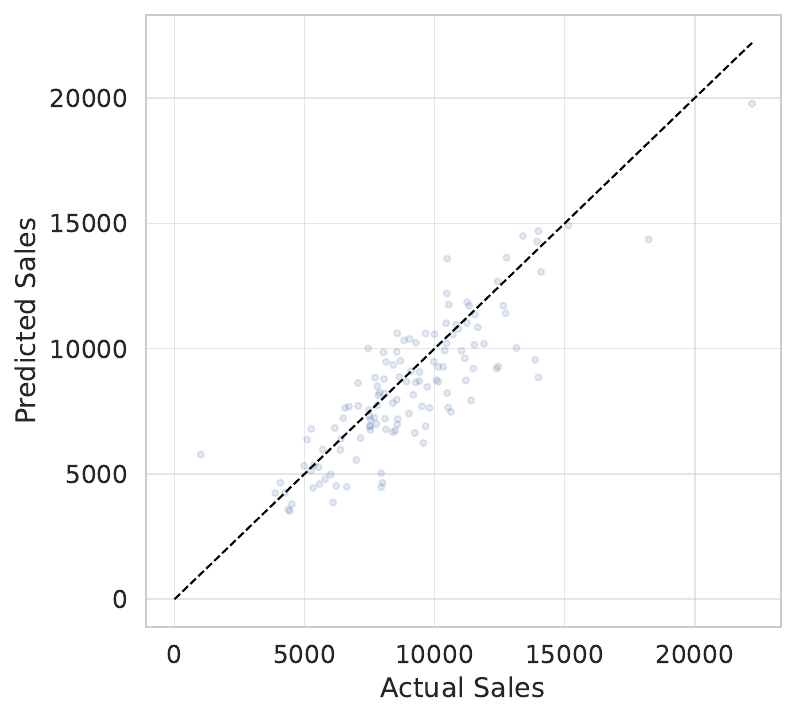}
    \caption{Predicted vs Actual Sales of the H4 model on the test set of the Rossmann dataset.}
    \label{fig:rossman_h4}
    \end{subfigure}
    
    \caption{Performance metrics of the adversarial training process for the H1 discriminator and predicted vs actual sales for the H4 model on the Rossmann dataset.}
    \label{fig:combined_fig_app}
\end{figure}

Figure \ref{fig:rossman_h4} presents the predicted versus actual sales of the H4 model on the test set of the Rossmann dataset.

The confusion matrix in Figure \ref{fig:disc_confusion_matrix} shows that while the discriminator does sometimes correctly guess the correct client and its behavior is not completely random, it still performs worse than random guessing on average.


\newpage
\section*{Paper Checklist}

\begin{enumerate}

\item {\bf Claims}
    \item[] Question: Do the main claims made in the abstract and introduction accurately reflect the paper's contributions and scope?
    \item[] Answer: \answerYes{} 
    \item[] Justification: We make targeted claims regarding the runtime, communication cost, and downstream utility of transformer models in private settings and demonstrate them with empirical results. We provide citations for claims we do not directly support with empirical evidence.

\item {\bf Limitations}
    \item[] Question: Does the paper discuss the limitations of the work performed by the authors?
    \item[] Answer: \answerYes{} 
    \item[] Justification: When describing our proposed approach in Section \ref{sec:methodology}, we clearly present the trade-off of each baseline and our approach, with the associated limitations. We also discuss in Section \ref{sec:prob_def} what trade-offs we are willing to make to achieve our goal.

\item {\bf Theory assumptions and proofs}
    \item[] Question: For each theoretical result, does the paper provide the full set of assumptions and a complete (and correct) proof?
    \item[] Answer: \answerNA{} 
    \item[] Justification: Our work relies on the privacy guarantees provided by MPC protocols. We do not propose new theory and instead propose a new approach.

    \item {\bf Experimental result reproducibility}
    \item[] Question: Does the paper fully disclose all the information needed to reproduce the main experimental results of the paper to the extent that it affects the main claims and/or conclusions of the paper (regardless of whether the code and data are provided or not)?
    \item[] Answer: \answerYes{} 
    \item[] Justification: We provide ample descriptions of our method, our baselines, the data we use, the experiment details, and our hyperparameters. From these, one could reproduce results similar to the ones we present in the paper.

\item {\bf Open access to data and code}
    \item[] Question: Does the paper provide open access to the data and code, with sufficient instructions to faithfully reproduce the main experimental results, as described in supplemental material?
    \item[] Answer: \answerYes{} 
    \item[] Justification: We provide the anonymized code for running our experiments in the supplementary material for our submission, along with an anonymized repository link containing the full codebase and model checkpoints in the appendix.

\item {\bf Experimental setting/details}
    \item[] Question: Does the paper specify all the training and test details (e.g., data splits, hyperparameters, how they were chosen, type of optimizer) necessary to understand the results?
    \item[] Answer: \answerYes{} 
    \item[] Justification: We provide the necessary details in the core of the paper, with further details in the appendix.

\item {\bf Experiment statistical significance}
    \item[] Question: Does the paper report error bars suitably and correctly defined or other appropriate information about the statistical significance of the experiments?
    \item[] Answer: \answerYes{} 
    \item[] Justification: We provide 95\% confidence intervals for our results where possible and relevant.

\item {\bf Experiments compute resources}
    \item[] Question: For each experiment, does the paper provide sufficient information on the computer resources (type of compute workers, memory, time of execution) needed to reproduce the experiments?
    \item[] Answer: \answerYes{} 
    \item[] Justification: We describe the exact hardware used to run our experiments in the Appendix.
    
\item {\bf Code of ethics}
    \item[] Question: Does the research conducted in the paper conform, in every respect, with the NeurIPS Code of Ethics \url{https://neurips.cc/public/EthicsGuidelines}?
    \item[] Answer: \answerYes{} 
    \item[] Justification: We conform in every respect with the NeurIPS Code of Ethics.

\item {\bf Broader impacts}
    \item[] Question: Does the paper discuss both potential positive societal impacts and negative societal impacts of the work performed?
    \item[] Answer: \answerNo{} 
    \item[] Justification: We do not address societal impact in our paper as we focus strictly on the computational and communication bottlenecks of distributed inference infrastructure. Furthermore, the fundamental nature of our work inherently promotes a positive societal impact, protecting user data privacy, and does not introduce direct paths to negative applications.
    
\item {\bf Safeguards}
    \item[] Question: Does the paper describe safeguards that have been put in place for responsible release of data or models that have a high risk for misuse (e.g., pre-trained language models, image generators, or scraped datasets)?
    \item[] Answer: \answerNA{} 
    \item[] Justification: Our models pose no risk of misuse, and the data they are trained on, which we use for our experiments, is publicly available on the Kaggle website.

\item {\bf Licenses for existing assets}
    \item[] Question: Are the creators or original owners of assets (e.g., code, data, models), used in the paper, properly credited and are the license and terms of use explicitly mentioned and properly respected?
    \item[] Answer: \answerYes{} 
    \item[] Justification: We cite the requested citation for the Rossmann dataset and provide a link to the original work. The Credit Card Transactions Fraud Detection Dataset is part of the public domain (CC0) and we provide a link to the original work.

\item {\bf New assets}
    \item[] Question: Are new assets introduced in the paper well documented and is the documentation provided alongside the assets?
    \item[] Answer: \answerYes{} 
    \item[] Justification: Our code is documented, with a README describing how to reproduce our results and a license.

\item {\bf Crowdsourcing and research with human subjects}
    \item[] Question: For crowdsourcing experiments and research with human subjects, does the paper include the full text of instructions given to participants and screenshots, if applicable, as well as details about compensation (if any)? 
    \item[] Answer: \answerNA{} 
    \item[] Justification: We did not crowdsource nor use any human subjects.

\item {\bf Institutional review board (IRB) approvals or equivalent for research with human subjects}
    \item[] Question: Does the paper describe potential risks incurred by study participants, whether such risks were disclosed to the subjects, and whether Institutional Review Board (IRB) approvals (or an equivalent approval/review based on the requirements of your country or institution) were obtained?
    \item[] Answer: \answerNA{} 
    \item[] Justification: We did not crowdsource nor use any human subjects.

\item {\bf Declaration of LLM usage}
    \item[] Question: Does the paper describe the usage of LLMs if it is an important, original, or non-standard component of the core methods in this research? Note that if the LLM is used only for writing, editing, or formatting purposes and does \emph{not} impact the core methodology, scientific rigor, or originality of the research, declaration is not required.
    \item[] Answer: \answerNA{} 
    \item[] Justification: LLMs usage is not an important, original, or non-standard component of the core methods in our research.

\end{enumerate}

\end{document}